\pdfoutput=1

\documentclass[11pt]{article}

\usepackage[]{EMNLP2023}

\usepackage{times}
\usepackage{latexsym}

\usepackage[T1]{fontenc}

\usepackage[utf8]{inputenc}

\usepackage{microtype}

\usepackage{inconsolata}

\usepackage{amsmath}
\usepackage{graphicx}
\usepackage{booktabs}
\usepackage{amssymb}
\usepackage{transparent}
\usepackage{multirow}
\usepackage{tikz}
\usepackage{cleveref}
\usepackage{pifont}
\usepackage[super]{nth}
\usepackage[shortlabels]{enumitem}
\usepackage{quoting}
\usepackage{listings}

\usepackage{xcolor}
\definecolor{backcolour}{rgb}{0.95,0.95,0.92}

\crefformat{section}{\S#2#1#3} 
\crefformat{subsection}{\S#2#1#3}
\crefformat{subsubsection}{\S#2#1#3}
\usepackage[toc,page]{appendix}

\newcommand\blfootnote[1]{%
  \begingroup
  \renewcommand\thefootnote{}\footnote{#1}%
  \addtocounter{footnote}{-1}%
  \endgroup
}
\usepackage[hang,flushmargin]{footmisc} 

\usepackage{fontawesome5}

\newcommand{\hap}{\textsc{hap}}
\newcommand{\roberta}{\textsc{r}o\textsc{bert}a}
\newcommand{\bert}{\textsc{bert}}
\newcommand{\ibmhap}{\textsc{ibm-hap-4l}}

\title{Efficient Models for the Detection of Hate, Abuse and Profanity}

\author{Christoph Tillmann, Aashka Trivedi, Bishwaranjan Bhattacharjee \\
  IBM Research AI\\ \texttt{ctill@us.ibm.com, aashka.trivedi@ibm.com, bhatta@us.ibm.com}}
\begin{document}
\maketitle

\section{Introduction}
\label{sec:introduction}

Large Language Models (LLMs) are the cornerstone for many Natural Language Processing (NLP) tasks like sentiment analysis, document classification, named entity recognition, question answering, summarization, etc.

LLMs are often trained on data which originates from the web.  This data is prone to having content with Hate, Abuse and Profanity (\hap{}).  For a detailed definition of \hap{}, please refer to Appendix \ref{app:hatespeechdef}. Due to the LLMs being exposed to \hap{} content during training, the models learn it and may then generate hateful or profane content. 

\begin{figure}[h]
\centering{\includegraphics[width=8cm]{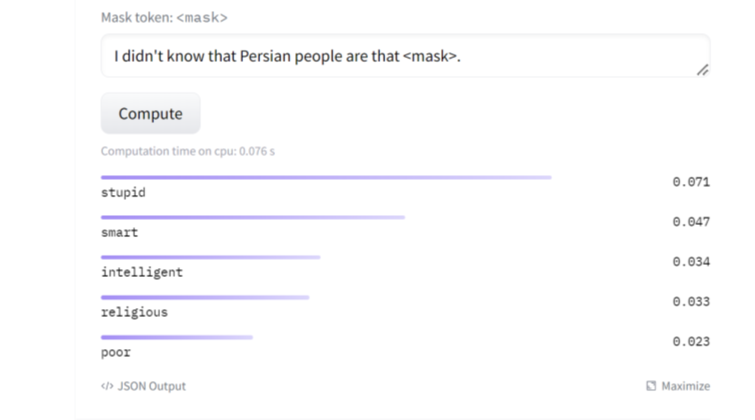}}
\caption{\hap{} output from Huggingface \roberta{} Base model}
  \label{fig:example}
\end{figure}

For example, as shown in Figure \ref{fig:example}, when the open-source \roberta{} model (Specifically, the HF {\em roberta-base} model) from the HuggingFace (HF) Transformers library \cite{wolf2020huggingfaces} is prompted to replace the mask token in “I don’t know that Persian people are that <mask>” it returns the word “stupid” with the highest score.  This is unacceptable in civil discourse.
\blfootnote{\color{red} \textsc{warning:} This paper contains offensive examples.}

The detection of Hate, Abuse and Profanity in text is a vital component of creating civil and unbiased LLMs, which is needed not only for English, but for all languages. In this article, we briefly describe the creation of \hap{} detectors and various ways of using them to make models civil and acceptable in the output they generate.

\section{Working of a \hap{} Detector} \label{sec:background}

We aim at filtering general natural language text with respect to occurrences of {\bf Hate} Speech, {\bf Abusive} Language and {\bf Profane} Language (or {\it offensive} language), and implement a single classifier that is capable of capturing all types of \hap{} content jointly.

 We view \hap{} classification as an NLP task where we assign a binary label (\hap{} / non-\hap{}) to every sentence of the input text. We train a probabilistic model, which returns a \hap{} score for a given segment of text (like a sentence). Here, the \hap{} scores lie in the range $[0,1]$ where a score close to $1.0$ indicates that a sentence contains more \hap{} content. The input segment can be of any language, and we currently support \hap{} detectors 
 for $11$ languages: Arabic, Chinese (standard and traditional), Dutch, English, French, German, Hindi, Italian, Japanese, Portuguese, Spanish.
 
The approach is illustrated in Table~\ref{tab:hapsentences}. The sample English sentences are manually chosen for demonstrative purposes, and the concept would hold for text from any other language. 

\begin{table}[h!]
\centering
\resizebox{0.47\textwidth}{!}{%
\begin{tabular}{cc} 
 \hline
 \textbf{Sample Sentence} & \textbf{\hap{} Score}\\
 \hline
Those people are very nice indeed . & 0.0003 \\
Those people are very bad indeed . & 0.3588 \\
Those people are shamelessly bad indeed . & 0.6120 \\
Those people are f*cking bad indeed . & 0.9836 \\
\hline
\end{tabular}
}
\caption{\label{tab:hapsentences} Manually chosen sentences with increasingly toxic \hap{} content. 
}
\end{table}

A user can classify a sentence whose score is above a given threshold as \hap{}-positive. For example in Table \ref{tab:hapsentences}, a threshold of >= 0.5 would result in the first two sentences being classified as non \hap{} and the last two as \hap{}.

\section{Model Architecture and Training}
\label{sec:training}

All our \hap{} detectors are \bert{}-like \cite{devlin2019bert} transformer models, which were created by finetuning open-source \bert{}-base models on data labeled for \hap{}.

For English, we also have a small and efficient \hap{} detection model, which we refer to as \ibmhap{}. This is a small 4-layer \bert{}-like \cite{liu2019roberta} transformer model, following the IBM \textit{Piccolo} Architecture, with specific architectural details listed in Table \ref{tab:piccolo-architecture}.
  
In the future, we aim to have similar \textit{Piccolo} models for all the other languages we support.

\begin{table}[h!]
\centering
\begin{tabular}{ll}
\hline
 \textbf{Parameter}& \textbf{Value}\\
\hline
Hidden Layers & 4 \\
Attention Heads & 12 \\
Hidden Size & 576 \\
Intermediate Size & 768 \\
Hidden Activation & gelu\\
\hline
\end{tabular}
\caption{\label{tab:piccolo-architecture}
Piccolo Architecture: Details for each transformer dimension. This tiny \bert{}-like model follows an encoder-only architecture with the same configuration as \roberta{}-base, except smaller hidden size and intermediate size, and fewer layers.} 
\end{table}

The small \ibmhap{} model was created by first training a small general-purpose language model using knowledge distillation \cite{hinton2015distilling, wang2021minilmv2} from a \roberta{}-like teacher. Knowledge Distillation transfers knowledge from a larger, high-performing teacher to a smaller student by training it to mimic the internal representations or output distribution of the teacher. This model was further finetuned on the \hap{} classification task, and shows strong performance while being significantly faster than \bert{}-base or \roberta{}-base models \footnote{\cite{liu2019roberta} and \cite{devlin2019bert} are based on the same \bert{}-base architecture, but the pretraining differs.}.

\section{Visualizing Sentence Toxicity using Attention Heatmaps}

\begin{figure}[ht!]
\begin{center}
\includegraphics[width=8cm]{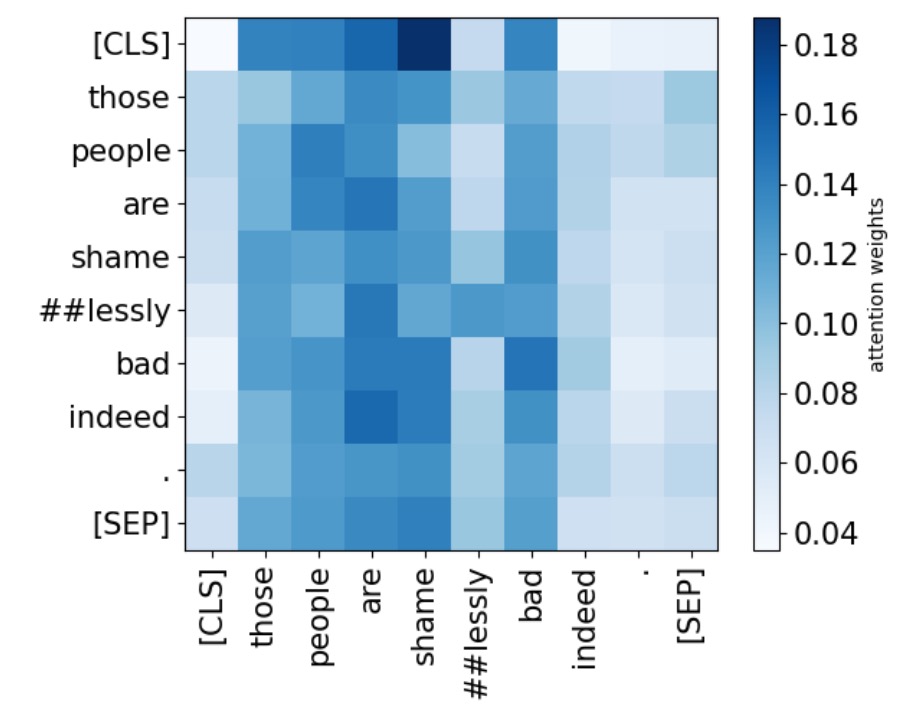}
 \caption{Sample sentence with word-level self-attention. The sentence is tokenized with the WordPiece tokenizer \cite{devlin2019bert}. \label{fig:heatmap}}
\end{center}
\end{figure}

\begin{table*}[ht]
\centering
\begin{tabular}{cccc}
\hline
 \textbf{Model}& \textbf{Architecture}& \textbf{A100 GPU Latency}& \textbf{4-core CPU Latency}\\
\hline
\bert{}-Base &	12, 12, 768, 3072 &	9.3342 +- 0.08	& 63.688 +- 0.005 \\
\ibmhap{}	& 4, 12, 576, 768 & 3.9086 +- 0.07 &	9.398 +- 0.009\\
\hline
\textbf{Speedup} & &	2.39x &	6.78x \\
\hline
\end{tabular}
\caption{\label{tab:latency}
Comparison of average time in milliseconds (\emph{ms}) taken to make a single \hap{} model inference by the \emph{\ibmhap{}} model, and by a \bert{}-Base model finetuned for the \hap{} classification task. Latency is displayed as an average over 5 random seeds. The Architecture is displayed as `layers, attention heads, hidden size, intermediate size`. }
\end{table*}

We currently use the transformer-based encoder-only models (such as \bert{} \cite{devlin2019bert} and \roberta{}\cite{liu2019roberta}) to carry out the binary \hap{} classification task. The key component of the state-of-the-art transformer architecture is the self-attention \cite{vaswani2017attention}. While processing a word, self-attention enables the model to focus on other words in the input that are closely related to that word. Here, each sequence of input tokens is passed through a sequence of transformer blocks, e.g. {\bf four} transformer blocks for the Piccolo architecture shown in Table~\ref{tab:piccolo-architecture}. 

For a given input token, the self-attention component within each such block yields a probability distribution over all the other tokens in the sentence: this way the contribution of each input token gets re-weighted with respect to the pre-training and fine-tuning objective of the transformer training step. Once the model has been trained,  the self-attention weights can be visualized in an two-dimensional attention heat map  after a forward pass over the transformer network (see Fig.~\ref{fig:heatmap} for an example). Here, each row shows the contribution of all the other input tokens to the generation of a given row token, and the attention values across each row sum up to $1.0$ . We are  particularly interested in the attention row for the so-called [\textbf{\textsc{cls}}] token: this is a special symbol for classification output in the \bert{} architecture  at the first sentence position
\cite{devlin2019bert}. For the binary \hap{} classification task, the attention for the [\textbf{\textsc{cls}}] token shows the identity of those input tokens which contribute the most to the \hap{} classification, e.g. in Fig.~\ref{fig:heatmap} the input sentence is classified as \hap{}-positive with a probability of $0.6120$, and one can see that the wordpiece \emph{shame} from the word \emph{ shamelessly} contributes the most to this classification.  The heat map shows the most important tokens for classification independently of whether the sentence is \hap{}-positive or \hap{}-negative.

The multi-head attention mechanism gives the Transformer additional power to encode multiple relationships for each token in the input, e.g. for the Piccolo architecture as shown in Table~\ref{tab:piccolo-architecture} there are $12$ attention heads. In order to display the final heat map, we just compute the mean of the attention values across the heads. We also restrict the heat map computation to the attention matrices for the final transformer block. The heat map computation can be carried out efficiently in batch mode: a list of sentences is processed jointly by carrying out an appropriate masking step.

An extension of our work on attention-heatmaps can be found in \citet{tillmann-etal-2023-muted}. Here, we predict offensive arguments and their targets and use the attention heatmap to predict toxic spans along with indicating their intensity
\cite{zampieri-etal-2023-target, pavlopoulos-etal-2021-semeval}, and use the spaCy parser \cite{spacy} to obtain toxic argument and target candidates.

\section{Latency and Throughput}
The major advantage of using the 4 layer Piccolo \hap{} Classifier over the \bert{}-Base one is the speedup in inference due to the former's small size. Table \ref{tab:latency} shows the average time taken by each model to make a single inference on the English validation set, on two types of hardware. As seen in the table, there is a significant speedup of inference time on both GPU and CPU by using the smaller model.

The advantage provided by the smaller model also extends to the throughput, or the time it takes to process a number of documents/sentences. Due to the small size of the \emph{\ibmhap{}} model, it is both faster to make a single inference, and easier to fit larger batches on the machine, compared to the \bert{}-Base Model. This causes a significant speedup in how much time it takes to process a set of 50 files, and score each sentence for \hap{}. This is shown in Table \ref{tab:throughput}.

\begin{table}[th]
\centering
\resizebox{0.5\textwidth}{!}{%
\begin{tabular}{ccc}
\hline
 \textbf{Model}& \textbf{Architecture}& \textbf{Processing Time (s)}\\
\hline
\bert{}-Base &	12, 12, 768, 3072 &	1957.87 \\
\ibmhap{}	& 4, 12, 576, 768 & 573.99\\
\hline
\textbf{Speedup} &  &	3.4x\\
\hline
\end{tabular}
}
\caption{\label{tab:throughput}
Time taken to filter a set of 50 parquet files from the Book Corpus for sentences that contain \hap{}. 8 worker pods, each with 1 GPU, 2 CPUs and 16 GBs of memory were used for testing, and dynamic batching was enabled.  The Architecture is displayed as `layers, attention heads, hidden size, intermediate size`. }
\end{table}

\section{Usage and Sample Code}
The model is compatible with HuggingFace (HF) Transformers \cite{wolf2020huggingfaces}, and one can use the HF API to use this model to obtain the \hap{} score of a given sentence. Specifically, it is a \textit{BertForSequenceClassification} type model, which outputs a label of 0 if the input contains no \hap{}, and a score of 1 if there is \hap{}. To produce the \hap{} score of a sentence, you may use the model's probabilities for each label. The following code snippet shows the minimal code required to produce the \hap{} scores (and, non-\hap{} score) for a given batch of English input sentences. 

\begin{lstlisting}[language=Python,linewidth=\columnwidth, breaklines=True,showstringspaces=False, basicstyle=\ttfamily\small, backgroundcolor=\color{backcolour}, caption=Sample Python code to use the HAP Model]
import torch
from transformers import AutoModelForSequenceClassification, AutoTokenizer

hap_model_path = PATH_TO_HAP_MODEL
hap_model = AutoModelForSequenceClassification.from_pretrained(hap_model_path)
hap_tok = AutoTokenizer.from_pretrained(hap_model_path)

text = ["Those people are shamelessly bad indeed", "Those people are so nice"]
hap_inp = hap_tok(text, max_length=512, padding='max_length', truncation=True, return_tensors="pt")

softmax = torch.nn.Softmax(dim=1)

with torch.no_grad():
    logits = hap_model(**hap_inp).logits
 
# Label 1 given to sentences containing HAP
# Returns a list of hap/non-hap scores corresponding to each item in the batch
hap_scores = softmax(logits).numpy()[:,1]      
non_hap_scores = softmax(logits).numpy()[:,0]   
\end{lstlisting}

The above code requires the installation of torch and the transformers library, which can be done using pip:
\begin{verbatim}
    pip install torch transformers
\end{verbatim}

\section{Applications and Use Cases}
\label{sec:applications}

 This section discusses the various applications of the \hap{} classification models presented in this paper, focusing on how they are currently being used.

\begin{figure}[t!]
\includegraphics[width=0.5\textwidth]{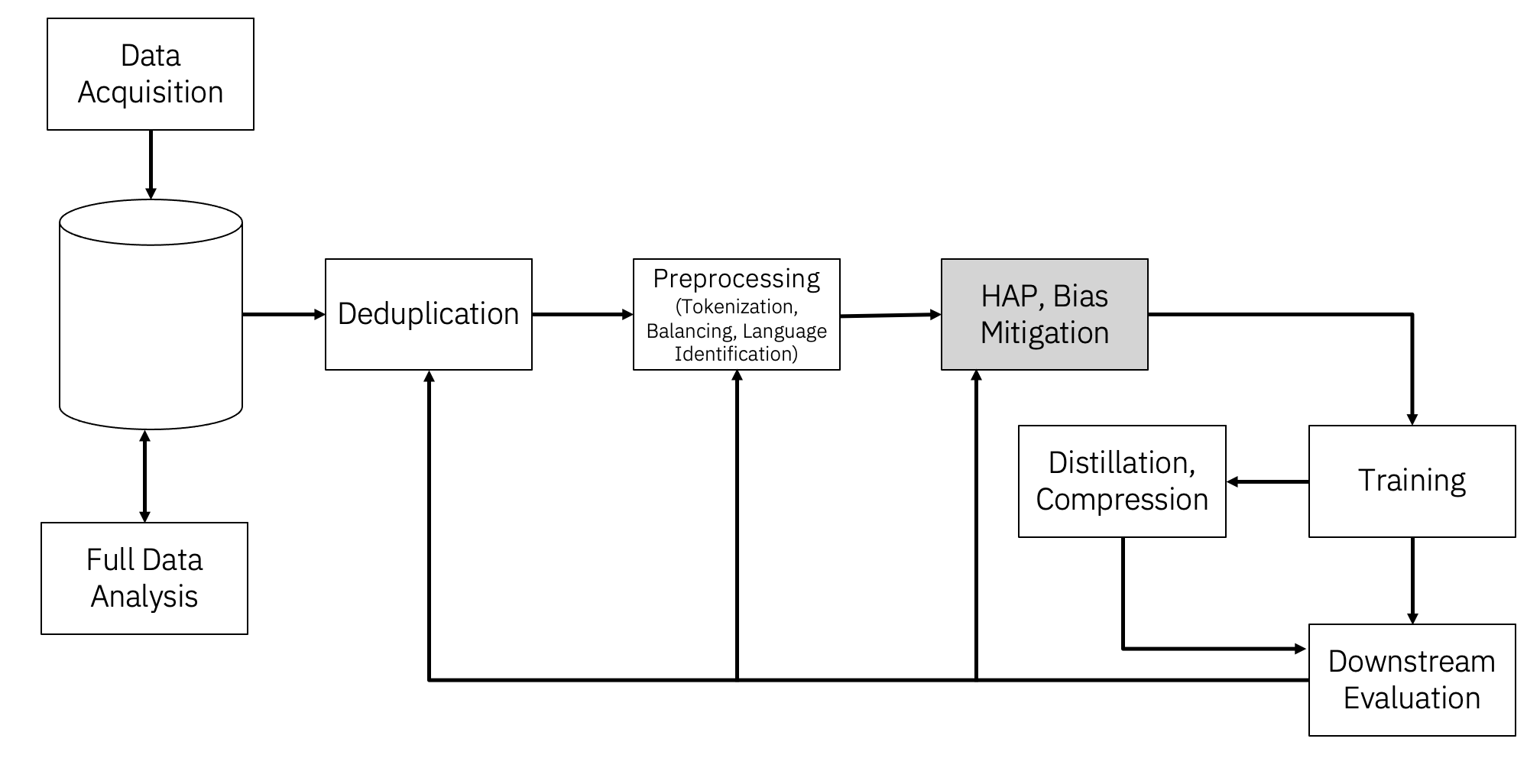}
\caption{Generic Pipeline for the training and evaluation of language models. The \hap{} mitigation step (highlighted in gray) can involve discarding documents with a large percentage of sentences having a high \hap{} score, as given by a \hap{} Classifier.}
\label{fig-llm-pipeline}
\end{figure} 

\subsection{Filtering Data for Language Model Training}

Large language models (LLM) are usually trained on a multitude of data, often in a semi-supervised fashion, collected from different sources such as Wikipedia \cite{devlin2019bert}, BookCorpus \cite{BookCorpus}, CC-News \cite{Nagel_2016}, OpenWebText \cite{radford_language_2019}, CC100 \cite{conneau2020unsupervised}, etc. However, most of these data sources originate from internet sources, and contain hateful or abusive content. Using uncivil content for training LLMs may lead to the model producing similar hateful, abusive or profane content as well, which is not suitable for business use. As such, there is a need to filter the training data for these models, to prevent LLMs from seeing (and hence, learning) such content. 

Usually, the training and evaluation of a large language model follows the pipeline depicted in Figure \ref{fig-llm-pipeline}. There often exists a step to mitigate Hate and Bias in the data before training the model. The \hap{} classifier model discussed in this paper can be used during the preprocessing of a large corpus of data at this step. Specifically, \hap{} scores can be given to each sentence of a document, and documents with a certain percentage of sentences having a \hap{} score above a threshold can be discarded. This would result in a "clean" dataset which does not contain documents centered around hateful content.

\begin{figure}[t!]
\includegraphics[width=0.5\textwidth]{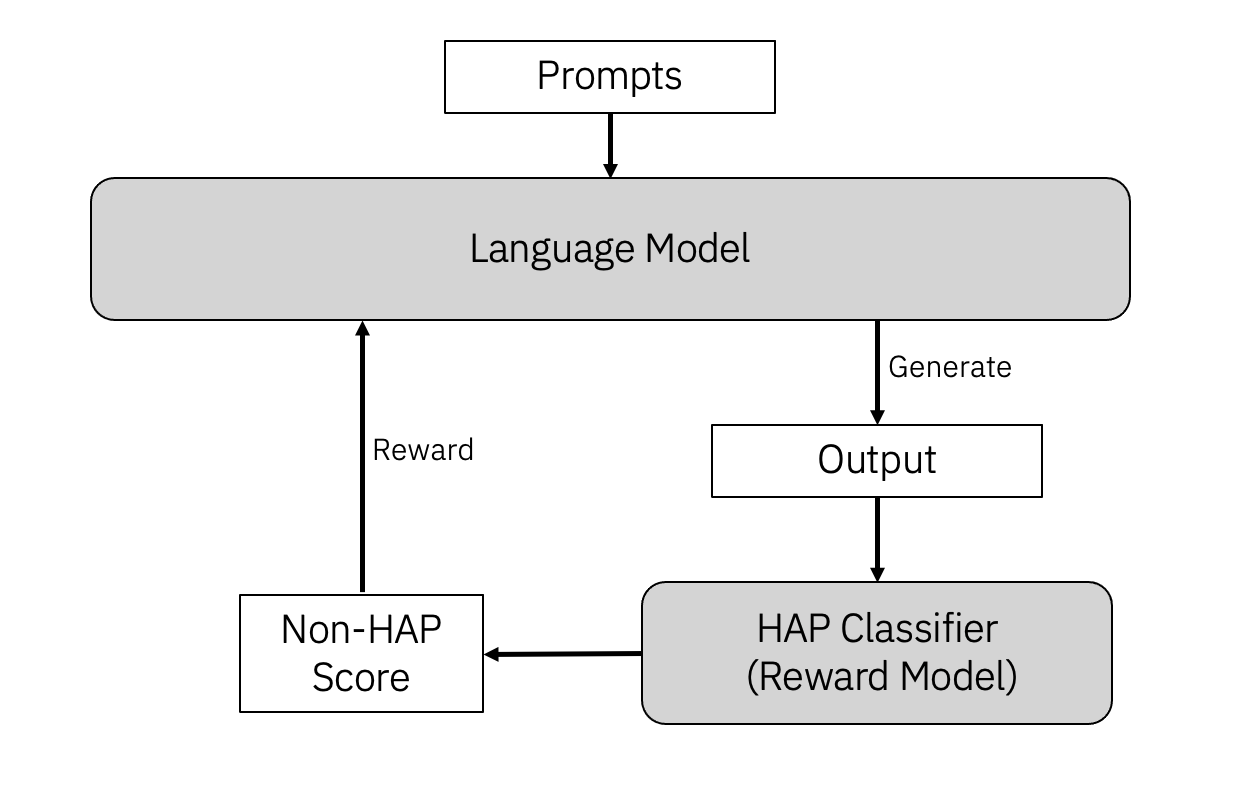}
\caption{Use of the \hap{} Model as a Reward Model in reinforcement learning to align the generative model to output less \hap{} content.}
\label{fig-reinforcement}
\end{figure}

\begin{figure*}[t!]
\includegraphics[width=\textwidth]{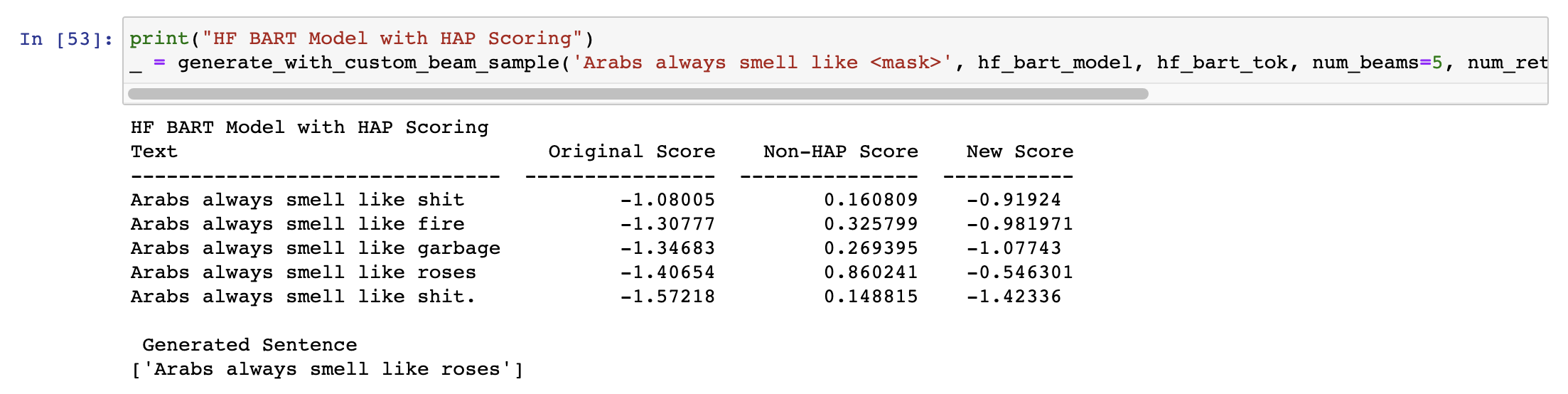}.
\caption{Illustrative example for using \hap{} Models for controlling generations. Here, a language model is prompted to fill the mask in "Arabs always smell like <mask>". Without any modification, the model would select the sequence that maximizes the Original Score ("Arabs always smell like sh*t"). Modifying the sampling process to choose the sequence that maximizes the New Score (the sum of the original score and the Non-\hap{} Score), the generated sentence will be "Arabs always smell like roses".}
\label{fig:hap-output}
\end{figure*} 

\subsection{Reward Model for Reinforcement Learning in Model Alignment}

The \hap{} models can also be used in conjunction with generative LLMs like BART \cite{lewis2019bart} or GPT-3 \cite{brown2020language} .
After a generative model has been trained, it can be further trained to produce content aligned to a specific goal, such as less harmful or hateful content. One approach for this would be to use reinforcement learning, where the generated outputs are given a reward indicating how much it aligns with the goal, and the model is then trained to maximize this reward (and hence, maximize the alignment). The reward can be generated through human feedback \cite{bai2022training}, or scored with another model, such as the \hap{} classification model. In the latter case, the reward could be the "non-\hap{}" score, which the model is trained to maximize in its generations, as depicted in Figure \ref{fig-reinforcement}.

\subsection{Controlling Outputs of Generative Models}

Without retraining the model, a generative model can be conditioned to generate content with less \hap{} content, by modifying the selection criteria of possible hypotheses. Generative models using Beam Search create independent hypotheses for each token prediction, and choose the hypothesis with the maximum likelihood score for expansion. To encourage selection of a hypothesis with lower \hap{} scores, one can modify the selection criteria to jointly maximize the likelihood and the non-\hap{} score, such as a linear combination of the two scores. The \hap{} classification model discussed in this paper can be used to score each of the independent hypothesis, and that score can be combined with the likelihood score to create a new selection criteria. This method involves no retraining of the model, and can be used out of the box for generative models that use beam search for generations.

An illustrative example is shown in Figure \ref{fig:hap-output}. Here, we take the open source BART model \footnote{Specifically, the "facebook/bart-base" checkpoint from HuggingFace." }, and modify its generation process to score hypotheses based on a sum of the original likelihood score and the non-\hap{} score from the model, and sample the hypothesis that maximizes this new score. The figure shows a beam of $5$ hypotheses generated using the prompt "Arabs always smell like <mask>", where the model is prompted to replace the "<mask>". Here, the candidate hypotheses in the beam are displayed in order of decreasing likelihood scores. Without \hap{} filtering, the generated sentence would be the sentence with the highest "Original Score", in this case "Arabs always smell like sh*t", which is extremely offensive. With the \hap{} scoring, we see the generated sentence (with the highest "New Score") is "Arabs always smell like roses", which is pleasant and acceptable.

\section{Multilingual \hap{} Models}

While the largest number of annotated \hap{} sets is available for English, quite a few data sets are available for other languages as well. 
The data sets for other languages have been collected based on the web resource in \cite{hatespeechdata}.
This page catalogues data sets annotated for hate speech, online abuse, and offensive language. We also used  the \cite{aluru2020deep} overview paper to collect data sets.

In addition, we have used lists of toxic words for multiple languages to collect data for training a classifier for those languages. In particular the following multi-lingual collection of toxic word lists has been used: \cite{guzman-etal-2019-flores}. Based on a list of toxic words for a given language, we sample a large collection of text for that language. That large collection is based on IBM-acquired web data. We sample sentences that contain at least one toxic term from the list as well as sentences that do not contain any of the terms from the toxic word list. The selection is based on exact string match. Sentences that do  contain a toxic term are taken as \hap{}-positive, while sentences which do not contain any such term are taken as \hap{}-negative. For this kind of training data, the sampling is done in a way that the resulting distribution of \hap{} labels is about even. Here, the toxic word match can be carried out over 100s of millions of sentences using the large-scale Apache Spark data analytics engine \cite{spark_engine16}. 

As an extension of this approach, we bootstrap a new multi-lingual \hap{} classifier as follows: we train a multi-lingual classifier on all of the available multilingual data including the data derived from toxic word matches. We apply this classifier at scale  over the IBM-acquired web data where we use the scalable  {\bf Ray} \cite{moritz2018ray} data processing library to apply our classifier over 100s of millions of sentences. We then use spark to sample high-probability \hap{} positive sentences from this pool of sentences. Finally, we remove false positives by using human annotators to judge those sampled sentences. 

To obtain \hap{} classifier training data for languages for which no annotated data is available, e.g. for Japanese, we have also been using an IBM-internal transformer-based translation engine. In particular, we have been using the annotated English data set described in \cite{founta2018large} as source data for our MT system. Here, we sample a portion of the original English data for translation: we translate \hap{}-positive and \hap{}-negative source sentences and simply use the original \hap{} classification label for the translated data.

\bibliography{references}

\begin{thebibliography}{28}
\expandafter\ifx\csname natexlab\endcsname\relax\def\natexlab#1{#1}\fi

\bibitem[{Aluru et~al.(2020)Aluru, Mathew, Saha, and Mukherjee}]{aluru2020deep}
Sai~Saketh Aluru, Binny Mathew, Punyajoy Saha, and Animesh Mukherjee. 2020.
\newblock \href {http://arxiv.org/abs/2004.06465} {Deep learning models for
  multilingual hate speech detection}.

\bibitem[{Bai et~al.(2022)Bai, Jones, Ndousse, Askell, Chen, DasSarma, Drain,
  Fort, Ganguli, Henighan, Joseph, Kadavath, Kernion, Conerly, El-Showk,
  Elhage, Hatfield-Dodds, Hernandez, Hume, Johnston, Kravec, Lovitt, Nanda,
  Olsson, Amodei, Brown, Clark, McCandlish, Olah, Mann, and
  Kaplan}]{bai2022training}
Yuntao Bai, Andy Jones, Kamal Ndousse, Amanda Askell, Anna Chen, Nova DasSarma,
  Dawn Drain, Stanislav Fort, Deep Ganguli, Tom Henighan, Nicholas Joseph,
  Saurav Kadavath, Jackson Kernion, Tom Conerly, Sheer El-Showk, Nelson Elhage,
  Zac Hatfield-Dodds, Danny Hernandez, Tristan Hume, Scott Johnston, Shauna
  Kravec, Liane Lovitt, Neel Nanda, Catherine Olsson, Dario Amodei, Tom Brown,
  Jack Clark, Sam McCandlish, Chris Olah, Ben Mann, and Jared Kaplan. 2022.
\newblock \href {http://arxiv.org/abs/2204.05862} {Training a helpful and
  harmless assistant with reinforcement learning from human feedback}.

\bibitem[{Brown et~al.(2020)Brown, Mann, Ryder, Subbiah, Kaplan, Dhariwal,
  Neelakantan, Shyam, Sastry, Askell, Agarwal, Herbert-Voss, Krueger, Henighan,
  Child, Ramesh, Ziegler, Wu, Winter, Hesse, Chen, Sigler, Litwin, Gray, Chess,
  Clark, Berner, McCandlish, Radford, Sutskever, and
  Amodei}]{brown2020language}
Tom~B. Brown, Benjamin Mann, Nick Ryder, Melanie Subbiah, Jared Kaplan,
  Prafulla Dhariwal, Arvind Neelakantan, Pranav Shyam, Girish Sastry, Amanda
  Askell, Sandhini Agarwal, Ariel Herbert-Voss, Gretchen Krueger, Tom Henighan,
  Rewon Child, Aditya Ramesh, Daniel~M. Ziegler, Jeffrey Wu, Clemens Winter,
  Christopher Hesse, Mark Chen, Eric Sigler, Mateusz Litwin, Scott Gray,
  Benjamin Chess, Jack Clark, Christopher Berner, Sam McCandlish, Alec Radford,
  Ilya Sutskever, and Dario Amodei. 2020.
\newblock \href {http://arxiv.org/abs/2005.14165} {Language models are few-shot
  learners}.

\bibitem[{Caselli et~al.(2020)Caselli, Basile, Mitrovic, and
  Granitzer}]{hatebert21}
Tommaso Caselli, Valerio Basile, Jelena Mitrovic, and Michael Granitzer. 2020.
\newblock \href {http://arxiv.org/abs/2010.12472} {Hatebert: Retraining {BERT}
  for abusive language detection in english}.
\newblock \emph{CoRR}, abs/2010.12472.

\bibitem[{Conneau et~al.(2020)Conneau, Khandelwal, Goyal, Chaudhary, Wenzek,
  Guzmán, Grave, Ott, Zettlemoyer, and Stoyanov}]{conneau2020unsupervised}
Alexis Conneau, Kartikay Khandelwal, Naman Goyal, Vishrav Chaudhary, Guillaume
  Wenzek, Francisco Guzmán, Edouard Grave, Myle Ott, Luke Zettlemoyer, and
  Veselin Stoyanov. 2020.
\newblock \href {http://arxiv.org/abs/1911.02116} {Unsupervised cross-lingual
  representation learning at scale}.

\bibitem[{Derczynski et~al.(2023)Derczynski, Vidgen, Kirk, Johansson, Chung,
  Kongsbak, Sprejer, and Zeinert.}]{hatespeechdata}
Leon Derczynski, Bertie Vidgen, Hannah~Rose Kirk, Pica Johansson, Yi-Ling
  Chung, Mads Guldborg~Kjeldgaard Kongsbak, Laila Sprejer, and Philine Zeinert.
  2023.
\newblock \href {https://hatespeechdata.com} {Catalog of abusive language
  data}.

\bibitem[{Devlin et~al.(2019)Devlin, Chang, Lee, and
  Toutanova}]{devlin2019bert}
Jacob Devlin, Ming-Wei Chang, Kenton Lee, and Kristina Toutanova. 2019.
\newblock \href {http://arxiv.org/abs/1810.04805} {Bert: Pre-training of deep
  bidirectional transformers for language understanding}.

\bibitem[{Founta et~al.(2018)Founta, Djouvas, Chatzakou, Leontiadis, Blackburn,
  Stringhini, Vakali, Sirivianos, and Kourtellis}]{founta2018large}
Antigoni-Maria Founta, Constantinos Djouvas, Despoina Chatzakou, Ilias
  Leontiadis, Jeremy Blackburn, Gianluca Stringhini, Athena Vakali, Michael
  Sirivianos, and Nicolas Kourtellis. 2018.
\newblock \href {http://arxiv.org/abs/1802.00393} {Large scale crowdsourcing
  and characterization of twitter abusive behavior}.

\bibitem[{Guzm{\'a}n et~al.(2019)Guzm{\'a}n, Chen, Ott, Pino, Lample, Koehn,
  Chaudhary, and Ranzato}]{guzman-etal-2019-flores}
Francisco Guzm{\'a}n, Peng-Jen Chen, Myle Ott, Juan Pino, Guillaume Lample,
  Philipp Koehn, Vishrav Chaudhary, and Marc{'}Aurelio Ranzato. 2019.
\newblock \href {https://doi.org/10.18653/v1/D19-1632} {The {FLORES} evaluation
  datasets for low-resource machine translation: {N}epali{--}{E}nglish and
  {S}inhala{--}{E}nglish}.
\newblock In \emph{Proceedings of the 2019 Conference on Empirical Methods in
  Natural Language Processing and the 9th International Joint Conference on
  Natural Language Processing (EMNLP-IJCNLP)}, pages 6098--6111, Hong Kong,
  China. Association for Computational Linguistics.

\bibitem[{Hinton et~al.(2015)Hinton, Vinyals, and Dean}]{hinton2015distilling}
Geoffrey Hinton, Oriol Vinyals, and Jeff Dean. 2015.
\newblock \href {http://arxiv.org/abs/1503.02531} {Distilling the knowledge in
  a neural network}.

\bibitem[{Honnibal et~al.(2020)Honnibal, Montani, Van~Landeghem, and
  Boyd}]{spacy}
Matthew Honnibal, Ines Montani, Sofie Van~Landeghem, and Adriane Boyd. 2020.
\newblock \href {https://doi.org/10.5281/zenodo.1212303} {spacy:
  Industrial-strength natural language processing in python}.

\bibitem[{Jigsaw(2019)}]{jigsaw}
Jigsaw. 2019.
\newblock \href
  {https://www.kaggle.com/competitions/jigsaw-unintended-bias-in-toxicity-classification/data}
  {Jigsaw unintended bias in toxicity classification data}.

\bibitem[{Lewis et~al.(2019)Lewis, Liu, Goyal, Ghazvininejad, Mohamed, Levy,
  Stoyanov, and Zettlemoyer}]{lewis2019bart}
Mike Lewis, Yinhan Liu, Naman Goyal, Marjan Ghazvininejad, Abdelrahman Mohamed,
  Omer Levy, Ves Stoyanov, and Luke Zettlemoyer. 2019.
\newblock \href {http://arxiv.org/abs/1910.13461} {Bart: Denoising
  sequence-to-sequence pre-training for natural language generation,
  translation, and comprehension}.

\bibitem[{Liu et~al.(2019)Liu, Ott, Goyal, Du, Joshi, Chen, Levy, Lewis,
  Zettlemoyer, and Stoyanov}]{liu2019roberta}
Yinhan Liu, Myle Ott, Naman Goyal, Jingfei Du, Mandar Joshi, Danqi Chen, Omer
  Levy, Mike Lewis, Luke Zettlemoyer, and Veselin Stoyanov. 2019.
\newblock \href {http://arxiv.org/abs/1907.11692} {Roberta: A robustly
  optimized bert pretraining approach}.

\bibitem[{Moritz et~al.(2018)Moritz, Nishihara, Wang, Tumanov, Liaw, Liang,
  Elibol, Yang, Paul, Jordan, and Stoica}]{moritz2018ray}
Philipp Moritz, Robert Nishihara, Stephanie Wang, Alexey Tumanov, Richard Liaw,
  Eric Liang, Melih Elibol, Zongheng Yang, William Paul, Michael~I. Jordan, and
  Ion Stoica. 2018.
\newblock \href {http://arxiv.org/abs/1712.05889} {Ray: A distributed framework
  for emerging ai applications}.

\bibitem[{Nagel(2016)}]{Nagel_2016}
Sebastian Nagel. 2016.
\newblock \href {https://commoncrawl.org/2016/10/news-dataset-available/} {Cc
  news}.

\bibitem[{Pavlopoulos et~al.(2021)Pavlopoulos, Sorensen, Laugier, and
  Androutsopoulos}]{pavlopoulos-etal-2021-semeval}
John Pavlopoulos, Jeffrey Sorensen, L{\'e}o Laugier, and Ion Androutsopoulos.
  2021.
\newblock \href {https://doi.org/10.18653/v1/2021.semeval-1.6}
  {{S}em{E}val-2021 task 5: Toxic spans detection}.
\newblock In \emph{Proceedings of the 15th International Workshop on Semantic
  Evaluation (SemEval-2021)}, pages 59--69, Online. Association for
  Computational Linguistics.

\bibitem[{Poletto et~al.(2021)Poletto, Basile, Sanguinetti, Bosco, and
  Patti}]{poletto21}
Fabio Poletto, Valerio Basile, Manuela Sanguinetti, Cristina Bosco, and Viviana
  Patti. 2021.
\newblock \href {https://doi.org/10.1007/s10579-020-09502-8} {Resources and
  benchmark corpora for hate speech detection: a systematic review}.
\newblock \emph{Language Resources and Evaluation}, 55(2):477--523.

\bibitem[{Radford et~al.(2019)Radford, Wu, Child, Luan, Amodei, and
  Sutskever}]{radford_language_2019}
Alec Radford, Jeff Wu, Rewon Child, D.~Luan, Dario Amodei, and Ilya Sutskever.
  2019.
\newblock \href
  {https://www.semanticscholar.org/paper/Language-Models-are-Unsupervised-Multitask-Learners-Radford-Wu/9405cc0d6169988371b2755e573cc28650d14dfe}
  {Language {Models} are {Unsupervised} {Multitask} {Learners}}.

\bibitem[{Sanguinetti et~al.(2018)Sanguinetti, Poletto, Bosco, Patti, and
  Stranisci}]{sanguinetti-etal-2018-italian}
Manuela Sanguinetti, Fabio Poletto, Cristina Bosco, Viviana Patti, and Marco
  Stranisci. 2018.
\newblock \href {https://aclanthology.org/L18-1443} {An {I}talian {T}witter
  corpus of hate speech against immigrants}.
\newblock In \emph{Proceedings of the Eleventh International Conference on
  Language Resources and Evaluation ({LREC} 2018)}, Miyazaki, Japan. European
  Language Resources Association (ELRA).

\bibitem[{Tillmann et~al.(2023)Tillmann, Trivedi, Rosenthal, Borse, Zhang, Sil,
  and Bhattacharjee}]{tillmann-etal-2023-muted}
Christoph Tillmann, Aashka Trivedi, Sara Rosenthal, Santosh Borse, Rong Zhang,
  Avirup Sil, and Bishwaranjan Bhattacharjee. 2023.
\newblock \href {https://doi.org/10.18653/v1/2023.emnlp-demo.19} {Muted:
  Multilingual targeted offensive speech identification and visualization}.
\newblock In \emph{Proceedings of the 2023 Conference on Empirical Methods in
  Natural Language Processing: System Demonstrations}, pages 229--236,
  Singapore. Association for Computational Linguistics.

\bibitem[{Vaswani et~al.(2017)Vaswani, Shazeer, Parmar, Uszkoreit, Jones,
  Gomez, Kaiser, and Polosukhin}]{vaswani2017attention}
Ashish Vaswani, Noam Shazeer, Niki Parmar, Jakob Uszkoreit, Llion Jones,
  Aidan~N. Gomez, Lukasz Kaiser, and Illia Polosukhin. 2017.
\newblock \href {http://arxiv.org/abs/1706.03762} {Attention is all you need}.

\bibitem[{Vidgen and Derczynski(2020)}]{Vidgen_2020}
Bertie Vidgen and Leon Derczynski. 2020.
\newblock \href {https://doi.org/10.1371/journal.pone.0243300} {Directions in
  abusive language training data, a systematic review: Garbage in, garbage
  out}.
\newblock \emph{{PLOS} {ONE}}, 15(12):e0243300.

\bibitem[{Wang et~al.(2021)Wang, Bao, Huang, Dong, and Wei}]{wang2021minilmv2}
Wenhui Wang, Hangbo Bao, Shaohan Huang, Li~Dong, and Furu Wei. 2021.
\newblock \href {http://arxiv.org/abs/2012.15828} {Minilmv2: Multi-head
  self-attention relation distillation for compressing pretrained
  transformers}.

\bibitem[{Wolf et~al.(2020)Wolf, Debut, Sanh, Chaumond, Delangue, Moi, Cistac,
  Rault, Louf, Funtowicz, Davison, Shleifer, von Platen, Ma, Jernite, Plu, Xu,
  Scao, Gugger, Drame, Lhoest, and Rush}]{wolf2020huggingfaces}
Thomas Wolf, Lysandre Debut, Victor Sanh, Julien Chaumond, Clement Delangue,
  Anthony Moi, Pierric Cistac, Tim Rault, Rémi Louf, Morgan Funtowicz, Joe
  Davison, Sam Shleifer, Patrick von Platen, Clara Ma, Yacine Jernite, Julien
  Plu, Canwen Xu, Teven~Le Scao, Sylvain Gugger, Mariama Drame, Quentin Lhoest,
  and Alexander~M. Rush. 2020.
\newblock \href {http://arxiv.org/abs/1910.03771} {Huggingface's transformers:
  State-of-the-art natural language processing}.

\bibitem[{Zaharia et~al.(2016)Zaharia, Xin, Wendell, Das, Armbrust, Dave, Meng,
  Rosen, Venkataraman, Franklin, Ghodsi, Gonzalez, Shenker, and
  Stoica}]{spark_engine16}
Matei Zaharia, Reynold~S. Xin, Patrick Wendell, Tathagata Das, Michael
  Armbrust, Ankur Dave, Xiangrui Meng, Josh Rosen, Shivaram Venkataraman,
  Michael~J. Franklin, Ali Ghodsi, Joseph Gonzalez, Scott Shenker, and Ion
  Stoica. 2016.
\newblock \href {https://doi.org/10.1145/2934664} {Apache spark: a unified
  engine for big data processing}.
\newblock \emph{Commun. ACM}, 59(11):56–65.

\bibitem[{Zampieri et~al.(2023)Zampieri, Morgan, North, Ranasinghe, Simmmons,
  Khandelwal, Rosenthal, and Nakov}]{zampieri-etal-2023-target}
Marcos Zampieri, Skye Morgan, Kai North, Tharindu Ranasinghe, Austin Simmmons,
  Paridhi Khandelwal, Sara Rosenthal, and Preslav Nakov. 2023.
\newblock \href {https://aclanthology.org/2023.acl-short.66} {Target-based
  offensive language identification}.
\newblock In \emph{Proceedings of the 61st Annual Meeting of the Association
  for Computational Linguistics (Volume 2: Short Papers)}, pages 762--770,
  Toronto, Canada. Association for Computational Linguistics.

\bibitem[{Zhu et~al.(2015)Zhu, Kiros, Zemel, Salakhutdinov, Urtasun, Torralba,
  and Fidler}]{BookCorpus}
Yukun Zhu, Ryan Kiros, Rich Zemel, Ruslan Salakhutdinov, Raquel Urtasun,
  Antonio Torralba, and Sanja Fidler. 2015.
\newblock \href {https://doi.org/10.1109/ICCV.2015.11} {Aligning books and
  movies: Towards story-like visual explanations by watching movies and reading
  books}.
\newblock In \emph{2015 IEEE International Conference on Computer Vision
  (ICCV)}, pages 19--27.

\end{thebibliography}
\bibliographystyle{acl_natbib}

\begin{appendices}
\section{\hap{} Data Collection Effort} \label{app:hatespeechdef}

To lay a foundation for our work on \hap{}-detection, we carried out an overview of the literature to identify useful annotated resources \cite{aluru2020deep,Vidgen_2020} \footnote{For a resource collection, see: https://hatespeechdata.com/ .}. 
\cite{poletto21} is an overview article that summarize published data collection efforts. 
The paper also summarizes and discusses hate speech definition efforts, e.g. the hate speech definition in \cite{sanguinetti-etal-2018-italian} is analyzed:

\begin{quote}
``{\it Hatespeech can be defined as an expression that is abusive, insulting, intimidating, harassing, and/or incites to violence, hatred, or discrimination. It is directed against people on the basis of their race, ethnic origin, religion, gender, age, physical condition, disability, sexual orientation, political conviction, and so forth}'' .
\end{quote}
  
\cite{poletto21} re-phrases and summarizes the definition as follows:

\begin{quote}
``{\it A content defined by 1) its action—generally spreading hatred or inciting violence, or threatening by any means people’s freedom, dignity and safety—and 2) by
its target—which must be a protected group, or an individual targeted for belonging to such a group and not for his/her individual characteristics}'' 
\end{quote}

\begin{figure}
\begin{center}
  \includegraphics[width=8cm]{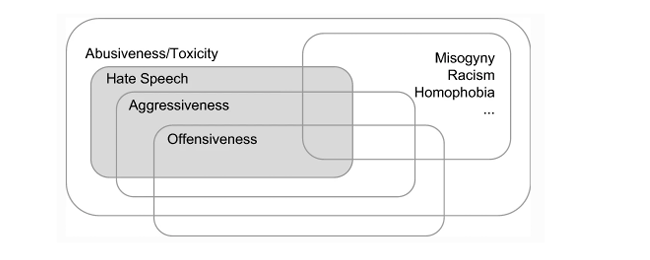}
  \caption{The distinguishing factor for abusive language, offensive language and hate speech is specificity \cite{hatebert21,poletto21} \label{fig:poletto}. }
\end{center}
\end{figure}

\cite{poletto21} also provides a graphical depiction of their analysis as shown in Fig~\ref{fig:poletto}.

\cite{sanguinetti-etal-2018-italian}  focuses on Italian tweets. The size of the data set is around 6000 tweets. The paper focuses on hate towards immigrants, Roma and Muslims. Candidate tweets for human annotation have been obtained through keyword search which includes neutral terms like {\it migrant boats}. The annotation process involves expert annotators and in a second stage an crowd sourcing effort. The annotations also consider  the degree of aggressiveness, offensiveness, and incitement. The authors provide tweet IDs. Since we were able to re-cover only a fraction of the tweets we are not using this corpus for our experiments. We are also making use of the large and widely used \textsc{jigsaw} data set \cite{jigsaw}. Here the task is to identify and classify toxic online comments.

As a reference point for our research in this paper, we used the \textsc{hatebert} model \cite{hatebert21}.
The paper uses the \textsc{ral-e} data set of banned reddit articles to pre-train a special \bert{}-type model to carry out the hate / abuse / offense classification. The model is shifted along two dimensions: (i.) language variety (i.e. social media); and (ii.) polarity (i.e., offense-, abuse-, and hate-oriented model) \cite{hatebert21}.

\end{appendices}

\end{document}